\documentclass[sigconf]{acmart}

\usepackage{natbib}
\usepackage{cuted}
\usepackage{tabularx,ragged2e}

\AtBeginDocument{%
  \providecommand\BibTeX{{%
    \normalfont B\kern-0.5em{\scshape i\kern-0.25em b}\kern-0.8em\TeX}}}

\setcopyright{acmlicensed}
\copyrightyear{2024}
\acmYear{2024}
\acmDOI{XXXXXXX.XXXXXXX}

\acmConference[Conference acronym 'XX]{Make sure to enter the correct
  conference title from your rights confirmation email}{June 03--05,
  2018}{Woodstock, NY}
\acmISBN{978-1-4503-XXXX-X/18/06}

\graphicspath{{./images/}} 

\usepackage{enumitem}
\usepackage{listings}
\usepackage{xcolor}
\usepackage{tcolorbox}

\begin{document}

\title{Evaluating the Retrieval Component in LLM-Based Question Answering Systems}

\author{Ashkan Alinejad}
\affiliation{%
  \institution{Thomson Reuters AI Labs}
  \city{Toronto}
  \country{Canada}
}
\email{ashkan.alinejad@thomsonreuters.com}

\author{Krtin Kumar}
\affiliation{%
  \institution{Thomson Reuters AI Labs}
  \city{Toronto}
  \country{Canada}
}
\email{krtin.kumar@thomsonreuters.com}

\author{Ali Vahdat}
\affiliation{%
  \institution{Thomson Reuters AI Labs}
  \city{Toronto}
  \country{Canada}
}
\email{ali.vahdat@thomsonreuters.com}






\renewcommand{\shortauthors}{Trovato and Tobin, et al.}

\begin{abstract}
  Question answering systems (QA) utilizing Large Language Models (LLMs) heavily depend on the retrieval component to provide them with domain-specific information and reduce the risk of generating inaccurate responses or hallucinations. Although the evaluation of retrievers dates back to the early research in Information Retrieval, assessing their performance within LLM-based chatbots remains a challenge.

  This study proposes a straightforward baseline for evaluating retrievers in Retrieval-Augmented Generation (RAG)-based chatbots. Our findings demonstrate that this evaluation framework provides a better image of how the retriever performs and is more aligned with the overall performance of the QA system. Although conventional metrics such as precision, recall, and F1 score may not fully capture LLMs' capabilities – as they can yield accurate responses despite imperfect retrievers – our method considers LLMs' strengths to ignore irrelevant contexts, as well as potential errors and hallucinations in their responses.
\end{abstract}




\received{20 February 2024}  
\received[revised]{12 March 2024} 
\received[accepted]{5 June 2024}  

\maketitle

\section{Introduction}

Recent advancements in Large Language Models (LLMs) \cite{llms} have shown promising results across a wide range of Natural Language Processing (NLP) tasks \cite{Open-LLM-Leaderboard-Report-2023}, including Information Retrieval (IR), text generation, and summarization. Particularly noteworthy is the significant improvement observed in Question Answering (QA) tasks, where the goal is to generate accurate responses given relevant document chunks containing the answer.

To enhance the accuracy of QA systems \cite{izacard-grave-2021-leveraging, rag} and mitigate the risk of hallucinations \cite{agrawal2024knowledge, shuster-etal-2021-retrieval-augmentation} from LLMs, Retrieval-Augmented Generation (RAG) models have been proved as a promising solution \cite{rag}. These models integrate a retriever component, which retrieves relevant document chunks to provide the LLM with the necessary context for generating responses. Evaluation of the retriever component typically relies on two types of metrics: (a) Rank-agnostic metrics, such as Precision and Recall, which compare retrieved chunks with gold-labeled chunks, and (b) Rank-aware metrics, such as Normalized Discounted Cumulative Gain (NDCG) \cite{ndcg} or Mean Reciprocal Rank (MRR) \cite{mrr}, which consider the order of retrieved documents.

Recent studies have proposed using LLMs as judges to evaluate various NLP tasks \cite{alpaca_eval, zheng2023judging, huang2024empirical} by providing them with an evaluation scheme in the form of a prompt. This approach enables LLMs to follow instructions and reliably evaluate responses based on user-defined metrics \cite{huang2024empirical}. However, while many studies have focused on using LLMs to evaluate QA responses, there is a notable gap in research concerning the evaluation of the retriever component \cite{salemi2024evaluating}.

In this paper, we introduce LLM-retEval, a framework designed to evaluate the retriever component in RAG-based question answering models. Our primary objective is to develop an evaluation metric for the retriever that considers the strengths and weaknesses of LLMs and provides a clearer understanding of its performance within an LLM-based QA system.

To achieve this goal, we first examine how conventional metrics for measuring retriever performance can fall short. Our experiments on the NQ-open corpus reveal that solely focusing on annotated data can significantly impact our ability to accurately assess retriever behavior, particularly when annotators fail to annotate all documents containing the answer. Additionally, evaluating the retriever as an isolated component overlooks the downstream effects of its results. This is crucial, as closely related but irrelevant retrieved chunks can easily distract LLMs from generating accurate responses.

Instead of solely examining the retriever's output, we propose measuring the performance of the downstream QA task relative to a model with an ideal retriever. By separately passing retrieved documents and gold documents to the answer generation LLM and comparing the resulting responses, we gain valuable insights into the retriever's effectiveness. Our findings suggest that by addressing the limitations of conventional metrics, they can become highly correlated with LLM-retEval, demonstrating the robustness of our method in capturing retriever performance.

\section{Evaluation Framework}
In this paper, we focus on evaluating the retriever in the QA task and we rely only on the correctness of the responses. The LLM-based QA system can be split into two distinct components. The retriever searches the corpus and extracts the documents that potentially contain the answer, and the generator receives those documents and generates an answer. We will start this section by formally describing the retriever and generator. Then we explain our evaluation method in Section \ref{subsec:method}.
\subsection{LLM-based Question Answering}
A Question Answering (QA) system tries to provide an accurate response to a natural language query from the user, based on relevant contexts from a provided pool of knowledge. Formally, Given a user query $\mathnormal{q}$, we try to extract the relevant information $\mathnormal{R}$ from the corpus of documents $\mathcal{D} = \{d_1, \dots, d_n\}$  to generate answer $\mathnormal{A}$. In an LLM-based QA system, this process is typically divided into two distinct components:

\subsection*{Retriever} Extracts a subset of documents $\mathnormal{R'}$ from $\mathnormal{D}$ which contains the answer to the query. Having an annotated set of relevant documents $\mathnormal{R}$, we traditionally try to maintain $\mathnormal{R'}$ as close as possible to $\mathnormal{R}$. Dense Passage Retrieval (DPR) \cite{karpukhin-etal-2020-dense} is a common method for extracting $R'$ which encodes the query and documents into the same vector space. The distance between the embeddings of the query and each document will be used to select $R'$.

\subsection*{Generator}
Attempts to generate an accurate response to the user query based on the extracted relevant documents ($\mathnormal{R'}$) by utilizing a language model $\mathcal{M}$:
$$ A = \mathcal{M}( \mathnormal{q}, \{\mathnormal{r}\} ),\ \ \ \ \ \forall{r} \in \mathnormal{R'} $$
The generator can receive the documents in $\mathnormal{R'}$ all at once, or it can receive them one by one or in batches depending on the document sizes, the complexity of the task, or the capabilities of the LLM.

\subsection{Evaluating retrieval in QA context}
\label{subsec:method}
In order to measure the performance of the retriever in the QA system, we pass its output ($R'$) to the generator LLM $\mathcal{M}$ to generate answers. We also pass the gold relevant documents ($R$) to the same LLM to see how the generator performs with an ideal retriever. By fixing the LLM parameters and comparing the answers from the two configurations, we can get a clear picture of how the retriever performs in an end-to-end question answering system. There are multiple methods for automatic comparison of the QA answers:

\begin{itemize}
    \item \textbf{Exact Match (EM)} compares strings directly to determine if they are exactly equal. However, it may be overly strict due to the potential variability in LLM outputs.
    \item \textbf{Token-based metrics} such as ROUGE-1, BLEU, and METEOR quantify the deviation between texts on a token/word level. Setting a threshold on these scores enables acceptance of answers that are highly similar but not exact matches.
    \item \textbf{Embedding-based metrics} vectorize answers and compute the cosine similarity between the vectors. BERTScore \cite{bertscore} is an example of such metric that is based on pretrained BERT embeddings which can capture the contextual information in answers.
    \item \textbf{LLM-based evaluation} have been recently utilized to evaluate the QA systems \cite{huang2024empirical, zheng2023judging}. They have demonstrated a great ability to capture the semantics of the answers while attending to their nuance variances.
\end{itemize}

We focus primarily on LLM-based evaluation to measure answer variances in this work. We can summarize our approach in three primary steps:

\begin{enumerate}
    \item Running the RAG-based QA system under evaluation. We use the retriever in our RAG pipeline to extract the relevant data and pass them to the generator LLM.
    \item Passing the gold relevant passages to the generator LLM to generate semi-gold responses.
    \item Comparing the responses generated by the RAG-based QA model with the semi-gold responses in step 2 using an LLM-based evaluation method, which outputs "Yes" if the responses match and "No" if they differ.
\end{enumerate}

It is important to note that the decision to use a yes/no grading system is based on the characteristics of the dataset. In the NQ-open dataset, questions are typically broad and the answers are short (fewer than five tokens.) However, when evaluating QA tasks in specialized domains such as legal or medical, where nuances in the answers are crucial, a more granular grading scale is recommended.

\section{Experimental setup}
In this section, we will describe our experimental settings, beginning with an explanation of the data used, followed by an outline of our evaluation methodology.

\subsection{Dataset}

For our experiments, we used the NQ-open dataset \cite{nqopen1, nqopen2}, which represents a subset of the Natural Questions (NQ) corpus \cite{nq}. Each sample in the NQ dataset includes a single question, a tokenized representation of the question, a Wikipedia URL, and the HTML representation of the corresponding Wikipedia page. The NQ-open dataset enhances this information by providing one or multiple answers for each question that are extracted directly from the associated Wikipedia passages and are no longer than 5 tokens. Although our evaluation approach doesn't require the gold answers to the queries, having them allows us to assess the model performance thoroughly and closely investigate how the retriever is performing relative to the overall QA performance.

Given the focus of our paper on evaluation criteria and the utilization of a basic embedding retrieval approach, we opt not to use the NQ-open training set. Instead, we only rely on the 2,889 samples in the test set for evaluation purposes. We use the English Wikipedia dump of 20 December 2018 for extracting the relevant chunks in our RAG model and follow the steps in \cite{cuconasu2024power} to split the passages into smaller chunks and clean the dataset. This gives us 21,035,236 documents to extract the relevant chunks.

\subsection{Retrieval and generative models}
In our experiments, we are utilizing a dense retrieval where we embed the document chunks using the "e5-large-v2" model\footnote{https://huggingface.co/intfloat/e5-large-v2} \cite{wang2022text}. We pick the top $k \in [ 1, 5, 10]$ documents based on the cosine similarity of the embeddings of query and chunks.

For the generation component of our QA models, we utilize two state-of-the-art language models: GPT-4 and ChatGPT-Turbo. These models have demonstrated strong performance in various natural language processing tasks, including question answering. Since for each question in NQ-open dataset, there might be multiple correct answers, we generate the ground-truth responses 3 times with a temperature of 0.5, to make sure that the variety of possible correct answers is generated.

For comparing the QA answers with the ground-truth model, we adopt the \textit{GPT4-Eval} \cite{kamalloo-etal-2023-evaluating} similar to the setup in \cite{adlakha2023evaluating}, which has been shown to be highly correlated with the human judgments. we have made slight modifications to the prompt to better align with the characteristics of our dataset. Details of these modifications are provided in Appendix \ref{sec:appendix}.

\section{Results}

Analyzing the performance of LLM-retEval starts by examining the failure cases of conventional metrics and LLM-retEval. We investigate the statistics of failure cases and assess the correlation between the two evaluation approaches. Although our experiments include reporting Precision@k and F1@k, our primary focus lies on comparing our method with Recall@k. This emphasis is due to the fact that in the NQ-open dataset, only one paragraph is labeled for each question. Consequently, the precision@k metric cannot exceed $1/k$ percent. In contrast, Recall@k indicates whether the correct answer appears within the top K results.

\subsection{Qualitative Analysis}

By closely examining cases that result in discrepancies between the conventional retriever metrics such as Precision or Recall and our LLM-retEval model, we gain insight into the limitations of traditional metrics in LLM-based QA models and how our proposed framework addresses these limitations. We identify these shortcomings through our experiments on the NQ-open dataset. It's important to note that testing additional datasets could reveal more failure cases.

\subsection*{\textbf{Failure cases of conventional metrics:}}

We categorize the sources of failure in traditional metrics into the following categories:

\begin{itemize}
    
    \item \textbf{Case A: Not annotating all the correct responses:} \\
    This occurs when an answer to a question can appear in multiple documents, but only one of them is labeled. This limitation is common in many databases where annotators are unable to search the entire corpus. As a result, traditional metrics penalize the retriever for not retrieving the gold excerpt. However, since the generator in LLM-based QA models can generate accurate responses using retrieved chunks, our model does not consider this scenario a failure.
    
    \item \textbf{Case B: Discrepancy between the searching documents and labeled data:} \\
     For example, consider the case illustrated in Figure \ref{figfail}, where the gold document is an older version of the same Wikipedia page being searched. In such cases, traditional metrics penalize the retriever for not returning the exact same chunk, even though its output is accurate and the generator can answer the question correctly based on context.
     
    \item \textbf{Case C: The retriever returns close but irrelevant chunks alongside the gold documents, distracting the generator:} This scenario is more common in LLM-based QA models. In such cases, the retriever receives a high score based on traditional metrics because it returns the gold documents. However, the presence of irrelevant chunks alongside the gold documents can lead the LLM to generate incorrect responses.
\end{itemize}

In the first example in Figure \ref{figfail}, the LLM correctly generates the answer \textit{Tulsa, Oklahoma} using a document other than the gold-labeled document. Here, the retriever effectively retrieves a document chunk containing the answer, although not the one labeled as correct by annotators.

The retrieved and gold documents in the second example in Figure \ref{figfail} come from the same Wikipedia page but their versions are different. Despite both containing the answer, the retriever is penalized by conventional metrics due to the discrepancy in indexing within our corpus. However, the generator successfully produces the correct response using the retrieved chunk.

In the final example illustrated in Figure \ref{figfail}, the retriever includes the correct document in its top 5 results. However, the addition of a misleading document about \textit{The Horn of Africa} distracts the generator, leading to an incorrect response (\textit{Nigeria, Horn of Africa}). While the retriever achieves a high Recall@5 by including the correct chunk, LLM-retEval penalizes it for including the misleading document. Notably, the low Precision@5 of 20 is due to the presence of only one correct chunk for each question, so the Precision@5 cannot go above 20.

\subsection*{\textbf{Failure cases of LLM-retEval:}}
There are three main sources of errors that come from inaccuracies in LLM responses:

\begin{itemize}
    \item \textbf{Case D: LLM cannot generate ground-truth answer}
    Generating an accurate response from the gold context is a crucial step in evaluating the retriever, and the LLMs can potentially fail to find the correct answer in the text. This usually happens when the information lies within a mal-processed table or text, or the answer to the question is not explicitly mentioned in the text.  

    \item \textbf{Case E: The ground-truth LLM doesn't generate all the correct answers:} This issue happens when there are multiple correct answers and the LLM returns a subset of them. For example, the answer to the question "who played scotty baldwins father on general hospital" can be both 'Peter Hansen', and 'Ross Elliott'. If the ground-truth LLM returns only one of them, then the other answer would be considered incorrect. This issue can be partly addressed by generating ground-truth responses multiple times with a temperature above 0.

    \item \textbf{Case F: Inaccuracies in comparing two answers} In this case the LLM is unable to precisely compare the generated response $R'$ with the ground-truth answer $R$.  This issue has been reported in previous studies \cite{kamalloo-etal-2023-evaluating, adlakha2023evaluating}. One possible solution to mitigate this problem is to integrate the LLM's judgments with alternative comparison techniques discussed in Section 2.2.
\end{itemize}

\begin{figure}[H]
    \centering
    \begin{tabular}{p{\columnwidth}}
    \hline
    \vspace{1mm}
    \textbf{Error Type:} Not annotating all the correct responses \\
    \textbf{Question:} where do the greasers live in the outsiders \\
    \textbf{Gold document ($R$):} The story in the book takes place in \textcolor{teal}{Tulsa, Oklahoma}, in 1965, but this is never explicitly stated in the book. \\
    \textbf{Retrieved document ($R'$):} In \textcolor{teal}{Tulsa, Oklahoma}, greasers are a gang of tough, low-income working - class teens. They include Ponyboy Curtis and his two older brothers, ... \\
    \textbf{Scores:} Recall@1: 0, Precision@1: 0, F1@1: 0, GPT4-retEval: Yes
    \vspace{1mm}\\
    \hline
    \vspace{1mm}
    \textbf{Error Type:} Discrepancy between the searching documents and labeled data \\
    \textbf{Question:} who got the first nobel prize in physics \\
    \textbf{Gold document ($R$):} The first Nobel Prize in Physics was awarded in 1901 to  \textcolor{teal}{Wilhelm Conrad Röntgen} ...  \textcolor{red}{As of 2017 , the prize has been awarded to 206 individuals}. There have been six years in which the Nobel Prize in Physics was not awarded ( 1916 , 1931 , 1934 , 1940 -- 1942 ) . \\
    \textbf{Retrieved document ($R'$):} The first Nobel Prize in Physics was awarded in 1901 to \textcolor{teal}{Wilhelm Conrad Röntgen} ... \textcolor{red}{As of 2016 , the prize has been awarded to 203 individuals} . There have been six years in which the Nobel Prize in Physics was not awarded ( 1916 , 1931 , 1934 , 1940 -- 1942 ) .\\
    \textbf{Scores:} Recall@1: 0, Precision@1: 0, F1@1: 0, GPT4-retEval: Yes
    \vspace{1mm} \\
    
    \hline

    \vspace{2mm}
    \textbf{Error Type:} The retriever returns close but irrelevant chunks alongside the gold documents, distracting the generator \\
    \textbf{Question:} in which regions are most of africa petroleum and natural gas found \\
    \textbf{Gold document ($R$):} \textcolor{teal}{Nigeria} is the largest oil and gas producer in Africa.  Crude oil from the \textcolor{teal}{delta basin} comes in ... \\
    \textbf{Gold answer:} Nigeria, delta basin \\
    \textbf{3rd retrieved document ($R'$):} \textcolor{red}{The Horn of Africa} is a peninsula in Northeast Africa . It juts hundreds of kilometers into ... \\
    \textbf{LLM answer based on top 5 docs:} Nigeria, Horn of Africa\\
    \textbf{Scores:} Recall@5: 100, Precision@5: 20, F1@5: 33.3, GPT4-retEval: No 
    \vspace{1mm}\\
    \hline
    \end{tabular}
    \vspace{-1mm}
    \caption{Qualitative examples cases where conventional metrics fail, along with LLM-retEval scores. The text in green color is the correct answer to the question.}
    \label{figfail}
\end{figure}



\begin{table}[t!]
\centering 
\begin{tabular}{|r|c|c|c|} 
\hline
 & @1 results & @5 results & @10 results \\ \hline \hline
 Recall failures & 500 & 248 & 161 \\ 
LLM-retEval Failures & 101 & 147 & 154 \\  \hline \hline

Recall@k & 0.498 & 0.792 & 0.865 \\  
Precision@k & 0.498 & 0.158 & 0.086 \\
F1@k & 0.498 & 0.264 & 0.157 \\ 
LLM-retEval & 0.648 & 0.767 & 0.786 \\ \hline \hline
GPT-Eval & 0.666 & 0.778 & 0.796 \\
\hline

\end{tabular}
\vspace{3mm}
\caption{The statistics of the failure cases along with the overall performance of the retriever and the QA model. Each "@k results" column presents the results of the model when we use the top k outputs of the retriever for generating answers.}
\label{tab:failurenumbers}
\end{table}

\subsection{Quantitative Analysis}
Table \ref{tab:failurenumbers} provides an overview of the failure cases, retriever performance, and overall QA results. Each column in the table represents how well the model performs when considering the top $k \in [1, 5, 10]$ results retrieved by the retriever. When we extract more chunks from the RAG pipeline, the likelihood of finding the correct answer in the retriever's output increases, leading to higher recall and a decrease in cases A and B. Although a higher value of $k$ may cause the LLM to become more easily distracted (case C), this increase is smaller than the decrease observed in cases A and B. As a result, overall failures in Recall decrease. However, increasing $k$ also raises the likelihood of the RAG-based QA model generating a response that is close but not entirely accurate. Thus, as we increase $k$ from 1 to 5, we observe Case F more frequently, leading to an increase in LLM-retEval failures. Nonetheless, overall, LLM-retEval maintains a consistently low failure rate across our experiments.

The second part of Table \ref{tab:failurenumbers} summarizes the retriever's performance based on conventional metrics and LLM-retEval. As $k$ increases, both precision and F1 scores drop significantly due to only one labeled chunk for each query. Therefore, our focus will be on recall. The last row of the table assesses how well the QA model performs based on the gold answers using the GPT-Eval model. Comparing the Recall@k and GPT-Eval results reveals that for lower values of $k$, the retriever's recall penalizes it for not leveraging the LLM's ability to generate correct responses from non-labeled chunks, while also giving undue credit to the retriever for not considering distractions from irrelevant retrieved chunks. Meanwhile, LLM-retEval consistently aligns closely with the overall QA performance across different numbers of retrieved chunks.

\begin{table*}[h!]
    \centering
    \begin{tabularx}{\columnwidth}{|c|cc|cc|cc|}
    \cline{1-7}
     & \multicolumn{2}{c|}{@1 results} & \multicolumn{2}{c|}{@5 results} & \multicolumn{2}{c|}{@10 results} \\ 
    \cline{2-7}
    Dataset        & ALL      & Refined  & ALL      & Refined & ALL      & Refined   \\ \cline{1-7}
    GPT4           & 0.58     & 0.87     & 0.48     & 0.80    & 0.44     & 0.73 \\
    chatGPT-turbo  & 0.55     & 0.80     & 0.48     & 0.78    & 0.43     & 0.71 \\ \cline{1-7}
    \end{tabularx}
    \vspace{3mm}
    \caption{Spearman's correlation of GPT models and Recall@$k$ when we retrieve $k \in [1, 5, 10]$ documents. The "ALL" dataset means NQ-open test set and "Refined" means the NQ-open test set after removing the Failure cases of conventional metrics.}
    \label{tab:correlation}
\end{table*}

We conducted Spearman's correlation \cite{Spearman} analysis between the LLM-retEval model and Recall@k, and the results are summarized in Table \ref{tab:correlation}. Correlation was measured on two versions of the test set: (1) "ALL," which includes all available data in our test set, and (2) "Refined," where failure cases of the recall are excluded. Due to the significant number of failure cases, Recall is not strongly correlated with LLM-retEval results when tested on the entire test set. However, upon removing these failure cases, the two evaluation approaches become highly correlated. With the increase in $k$, the correlation on the Refined dataset slightly decreases. This is mainly due to the fact that the Recall's failure cases decrease, leaving harder samples in the "Refined" set.

Table \ref{tab:correlation} also presents a comparison of the results between GPT4 and chatGPT-turbo. Both models exhibit very similar performance and are highly correlated with the "Refined" test set. However, GPT4 consistently maintains a closer correlation with Recall across our experiments, indicating that it experiences fewer failure cases compared to chatGPT-turbo. This suggests that GPT4 may be more adept at handling challenging scenarios and maintaining accuracy in retrieving relevant information.

\section{Related work}
Most evaluations of the RAG systems rely on metrics such as Precision or Recall \cite{books/aw/Baeza-YatesR99, alaofi2024generative}. These metrics treat the retriever as a standalone search engine and assess its performance independently of the downstream task. For systems outputting a ranked list, rank-aware metrics like Normalized Discounted Cumulative Gain (NDCG), Mean Reciprocal Rank (MRR), or Mean Average Precision (MAP) are commonly employed. While these metrics utilize human-annotated data to evaluate retriever performance, they overlook how the retriever's output influences the performance of subsequent components in the QA system.

Alternative approaches to measuring retriever behavior without relying on gold relevance judgments have been explored \cite{10.1145/383952.383961, 10.1145/383952.383964, li_automatic_performance_comparison}. These methods automate precision evaluation using techniques such as the vector space model and statistical analysis of documents. However, similar to conventional metrics like Precision or Recall, they fail to account for the downstream components of the QA pipeline, focusing solely on corpus statistics.

A related line of research investigates the use of Large Language Models (LLMs) to evaluate various NLP tasks \cite{alpaca_eval, zheng2023judging, huang2024empirical}. These studies emphasize end-to-end evaluation of LLM-based systems but do not provide clear insights into the impact of the underlying retriever component on performance.

A recent study by Salemi et al. (2024) \cite{salemi2024evaluating} explores the use of LLMs to measure retrieval quality in RAG-based models. Similar to our approach, they consider the downstream task when evaluating the retriever. However, their focus lies on enhancing the accuracy of conventional metrics such as Precision and Recall by leveraging LLMs to annotate relevance judgments for each question.

\section{Conclusion}
This paper investigates methods for evaluating retrievers in RAG-based question answering models and introduces a new metric for end-to-end performance assessment of the retriever. Our findings suggest that a direct comparison between the retrieved document set and the gold-relevant documents may not fully illustrate the performance of the retriever in LLM-based QA models. We discovered that generating answers based on gold-relevant documents using the same generator LLM can provide a reliable indicator of the retriever's performance. Our results demonstrate that this method is less susceptible to errors in LLM-based QA systems and aligns more closely with conventional metrics when failure cases are not considered. Furthermore, our observations reveal that while ChatGPT-Turbo demonstrates comparable performance to the GPT-4 model, the GPT-4 model consistently outperforms other models.




\bibliographystyle{ACM-Reference-Format}
\bibliography{references.bib}

\appendix

\newpage
\section{Prompts}
\label{sec:appendix}

\begin{figure}[h]
\begin{tcolorbox}[colback=white!10, boxrule=0pt, sharp corners, center]
\begin{lstlisting}[language=bash, basicstyle=\small\ttfamily, breaklines=true, breakindent=0em, columns=fullflexible]
Please read the question provided below and then review the accompanying document excerpts. Your task is to answer the question using the information from the documents:
Question: {question}

Relevant Document chunks:
{context}

After considering the information in the documents, please provide an answer (maximum 5 tokens) to the question: {question}.
Answer:
\end{lstlisting}
\end{tcolorbox}
\caption{The prompt template used for answer generation.}
\end{figure}

\begin{figure}[h]
\begin{tcolorbox}[colback=white!10, boxrule=0pt, sharp corners, center]
\begin{lstlisting}[language=bash, basicstyle=\small\ttfamily, breaklines=true, breakindent=0em, columns=fullflexible]
You are CompareGPT, a machine to verify the correctness of predictions. Answer with only "Yes" or "No". 
You are given a question, one or more corresponding ground-truth answers, and a prediction from a model. Compare the "Ground-truth answers" and the "Prediction" to determine whether the prediction correctly answers the question based on any of the provided ground-truth answers. 
All information in at least one of the ground-truth answers must be present in the prediction, including numbers and dates. You must answer "No" if the prediction does not completely match at least one set of specific details in the ground-truth answers. 
There should be no contradicting statements in the prediction. The prediction may contain extra information that does not contradict the ground-truth answers.

Question: {query} 
Ground-truth answers: {answer}
Prediction: {result}

Answer "Yes" if the prediction correctly answers the question based on any of the Ground-truth answers, otherwise answer "No"..
\end{lstlisting}
\end{tcolorbox}
\caption{The prompt template based on GPT-Eval for comparing LLM responses.}
\end{figure}

\end{document}